\documentclass[letterpaper]{article} % DO NOT CHANGE THIS
\usepackage{aaai25}  % DO NOT CHANGE THIS
\usepackage{times}  % DO NOT CHANGE THIS
\usepackage{helvet}  % DO NOT CHANGE THIS
\usepackage{courier}  % DO NOT CHANGE THIS
\usepackage[hyphens]{url}  % DO NOT CHANGE THIS
\usepackage{graphicx} % DO NOT CHANGE THIS
\usepackage{natbib}  % DO NOT CHANGE THIS AND DO NOT ADD ANY OPTIONS TO IT
\usepackage{caption} % DO NOT CHANGE THIS AND DO NOT ADD ANY OPTIONS TO IT
\usepackage{amsmath}
\usepackage{subfigure}
\usepackage{subcaption}
\usepackage{algorithm}
\usepackage{algorithmic}
\usepackage{color,soul}
\usepackage{comment}
\usepackage{newfloat}
\usepackage{listings}
\DeclareCaptionStyle{ruled}{labelfont=normalfont,labelsep=colon,strut=off} % DO NOT CHANGE THIS
\floatstyle{ruled}
\newfloat{listing}{tb}{lst}{}
\floatname{listing}{Listing}
\usepackage{multirow}
\usepackage{booktabs}

\title{Demystifying the Accuracy-Interpretability Trade-Off:\\ A Case Study of Inferring Ratings from Reviews}
\author {
    Pranjal Atrey,
    Michael P. Brundage,
    Min Wu,
    Sanghamitra Dutta
}
\affiliations {
    University of Maryland, College Park\\
    \{patrey,mbrundage,minwu,sanghamd\}@umd.edu
    }

\usepackage{bibentry}
\begin{document}

\maketitle

\begin{abstract}
Interpretable machine learning models offer understandable reasoning behind their decision-making process, though they may not always match the performance of their black-box counterparts. This trade-off between interpretability and model performance has sparked discussions around the deployment of AI, particularly in critical applications where knowing the rationale of decision-making is essential for trust and accountability. In this study, we conduct a comparative analysis of several black-box and interpretable models, focusing on a specific NLP use case that has received limited attention: inferring ratings from reviews. Through this use case, we explore the intricate relationship between the performance and interpretability of different models. We introduce a quantitative score called \emph{Composite Interpretability (CI)} to help visualize the trade-off between interpretability and performance, particularly in the case of composite models. Our results indicate that, in general, the learning performance improves as interpretability decreases, but this relationship is not strictly monotonic, and there are instances where interpretable models are more advantageous.
\end{abstract}


\section{Introduction}
Machine learning models now determine opportunities, from employment to credit, and directly shape public experiences, from classrooms to courtrooms, in ways that profoundly impact people's lives~\citep{dixon2020machine,nietzel2022colleges,barocas2016big,habehh2021machine}. To this end, black-box models have gained tremendous popularity because of their ability to discover complex patterns in data, e.g., Large Language Models (LLMs)~\citep{zhao2023survey} have surprised us with their remarkable potential at several natural language processing (NLP) tasks and beyond. Powerful, black-box models often operate in ways that are opaque to human stakeholders, raising concerns about their accountability and trust in critical decision-making~\citep{rudin2019stop,lipton2018mythos,lakkaraju2016interpretable,mishra2021survey,pmlr-v162-dutta22a,hamman2023robust,hamman2024robust,dissanayake2024model}. In fact, several existing regulations~\citep{gdpr} have strongly advocated for explainability so that people have the right to be informed about the logical reasoning underlying the decision-making process.

In recent years, the debate between black-box models and \emph{inherently} interpretable models has gained significant traction, with the ever-growing use of AI in several high-stakes applications. Interpretable models, such as decision trees and linear regression, offer more straightforward insights into the decision-making process, making them transparent and easier for humans to understand and trust~\citep{doshi2017towards, lakkaraju2016interpretable,letham2015interpretable,rudin2019stop,lipton2018mythos,mishra2021survey}. In contrast, black-box models, such as data-driven deep neural networks and LLMs, provide high performance for specific tasks but lack transparency, making it challenging to understand how they reach their conclusions. Recent research shows that black-box models are not universally superior; interpretable models can outperform them in certain applications, such as fault diagnosis or environmental predictions~\citep{liao2024explainable, ghasemi2023comparative}. \citet{rudin2019stop} argues that black-box models often lead to unreliable outcomes in high-stakes decisions, advocating for inherently interpretable models that improve transparency and trust. Similarly, \citet{li2020ib} propose IB-M, a flexible framework that combines the strengths of interpretable and black-box models to enhance decision-making. Striking a right balance between model performance and interpretability remains a key challenge, especially for critical, deployable AI applications where understanding decisions is essential. 

\noindent \textit{Main Contributions:} Towards addressing this challenge, in this work, we analyze the trade-off between interpretability and model performance for a specific NLP use case that has received limited attention: predicting product ratings from reviews. To demystify the trade-off between interpretability and model performance, we first introduce a quantitative score of interpretability, called \emph{Composite Interpretability (CI)} to allow us to rank different models based on their interpretability, particularly composite models that consist of different modules. Our scoring incorporates expert assessments of the simplicity, transparency, and explainability of machine learning models, while also factoring in model complexity by considering the number of parameters relative to the datasets. \emph{This ranking provides a pathway for ordering models by their degree of interpretability, extending beyond the traditional dichotomy of glass-box versus black-box models.} While prior research has predominantly addressed various classification models, our approach is particularly suited for composite models. Our results indicate a general trend where model performance improves as interpretability decreases, though this relationship is not strictly monotonic. Interestingly, we observe instances where interpretable models perform better than their black-box counterparts. This nuanced view underscores the importance of considering both interpretability and performance when selecting the most appropriate model for specific applications.

\section{Experimental Setup}
\subsection{Datasets}

Our datasets are retrieved from a larger database consisting of a large crawl of product reviews from Amazon \citep{ni2019justifying}. The database contains 82.83 million unique reviews from approximately 20 million users. The reviews are in text format while the ratings are in numerical format ranging from 1 to 5. To analyze product reviews and ratings, 40,000 product reviews and ratings have been extracted from the following four product categories leading to four datasets: (i) Cell Phones and Accessories (\textbf{CPA}); (ii) Office Products (\textbf{OP}); (iii) Automotive (\textbf{AM}); and (iv) Video Games (\textbf{VG}). To ensure a balanced dataset, each category consists of 2,000 reviews and ratings grouped by each rating star (1-5), consisting of a total of 10,000 reviews. Prior to experiments, the datasets were cleaned. Stop words such as \textit{the} and \textit{and} have been removed from the reviews to capture the more important words, and punctuation marks have also been removed. Using four different categories enables investigation of the generalizability of the results. 

\subsection{Different Types of Classification Models}

To differentiate between black-box and inherently interpretable models, we evaluate a range of models with varying levels of interpretability. 

VADER is a lexicon and rule-based sentiment analysis method tailored for social media expressions \citep{hutto2014vader}. VADER sentiment scores range between -1 and 1, where 1 represents positive text and -1 represents negative text. To keep the range of ratings and sentiment scores the same, scaling is used to convert the VADER sentiment scores to a range between 1 and 5 to predict ratings.

Next, for non-rule-based or learning-based models, we explore multiple embedding methods and classification models. Specifically, we deploy three embedding methods, ordered from most to least interpretable based on common perceptions: (i) CountVectorizer~\citep{pedregosa2011scikit}; (ii) Term Frequency-Inverse Document Frequency (TF-IDF)~\citep{brown1990statistical}; and (iii) Word2Vec~\citep{mikolov2013distributed}. We use four classification models, ordered from most to least interpretable, based on expert assessments: (i) logistic regression (LR)~\citep{hosmer2013applied}; (ii) multinomial naive Bayes (NB)~\citep{mccallum1998comparison}; (iii) kernel-based support vector machines (SVM)~\citep{cortes1995support}; and (iv) neural networks (NN)~\citep{goodfellow2016deep}. The neural network architecture used in our study is trained using a learning rate of $1e-4$ with two hidden layers of size 512 and 128, respectively, with ReLu activations embedded in between. The output layer is of size 5 and uses  \textit{softmax} activation. The training is executed using Adam optimizer. % with 20 epochs.

In addition to the purely rule-based and purely learning-based models, we also use a pre-trained language model fine-tuned for sentiment analysis: Bidirectional Encoder Representations from Transformers (BERT)~\citep{devlin2018bert}. Specifically, we utilize the HuggingFace \textit{bert-base-multilingual-uncasedmodel}, which has been fine-tuned for sentiment analysis on product reviews. This model predicts the sentiment of a review on a scale from 1 to 5. To deepen our understanding, we examine composite models by incorporating BERT sentiment scores as an additional input to the review embeddings for the classification task. This results in the following model configurations: (i) logistic regression with BERT sentiment scores (LR-BS); (ii) naive Bayes with BERT sentiment scores (NB-BS); (iii) support vector machines with BERT sentiment scores (SVM-BS); and (iv) neural networks with BERT sentiment scores (NN-BS).

\subsection{Proposed Composite Interpretability (CI) Score}

We propose a methodology for analyzing and visualizing the trade-off between interpretability and accuracy, termed Composite Interpretability (CI). This framework quantifies interpretability by incorporating multiple factors: simplicity, transparency, explainability, and complexity, as measured by the number of parameters. This approach combines expert assessments with quantitative metrics to comprehensively evaluate the multifaceted nature of interpretability in machine learning systems.

\begin{table}[!h]
\setlength{\tabcolsep}{1mm}
\centering
\caption{Ranking system for interpretability scores}
\small
\begin{tabular}{lccccc}
\hline
Model & Simp. & Transp. & Expl. & \# Params & \textbf{Interp.} \\
(Type) & (w=0.2) & (w=0.2) & (w=0.2) & (w=0.4) & \textbf{Score} \\
\hline
VADER & 1.45 & 1.60 & 1.55 & 0 & 0.20 \\
LR    & 1.55 & 1.70 & 1.55 & 3 & 0.22 \\
NB    & 2.30 & 2.55 & 2.60 & 15 & 0.35 \\
SVM   & 3.10 & 3.15 & 3.25 & 20,131 & 0.45 \\
NN    & 4.00 & 4.00 & 4.20 & 67,845 & 0.57 \\
BERT  & 4.60 & 4.40 & 4.50 & 183.7M & 1.00 \\
\hline
\end{tabular}
\label{tab:interpretabilityscores}
\end{table}

We collect rankings from 20 domain experts, who evaluate each model based on three fundamental interpretability criteria: simplicity, transparency, and explainability. Simplicity refers to the straightforwardness of the model's structure; transparency describes the ease of understanding the model's internal workings; and explainability reflects how effectively the model's predictions can be justified. 

Each model is ranked on a scale from 1 (most interpretable) to 5 (least interpretable) for each of the three criteria. The average rankings provided by 20 domain experts, along with the assigned weights (denoted as w) reflecting the relative importance of each criterion in the context of interpretability, are summarized in Table \ref{tab:interpretabilityscores}. The interpretability score of an individual model is calculated as follows:
\begin{equation}
\text{IS} = \sum_{c=1}^{C} \left( \frac{R_{m,c}}{R_{\text{max},c}} \cdot w_c \right) + \left( \frac{P_m}{P_{\text{max}}} \cdot w_{\text{parm}} \right)
\end{equation}
\noindent where: IS represents the interpretability score of the model; \(R_{m,c}\) is the ranking of model \(m\) for interpretability criterion \(c\); \( R_{\text{max}, c} \) is the maximum ranking for criterion \(c\) across all models; \(C\) is the total number of criteria ranked by domain experts; \(P_m\) is the number of parameters in model \(m\) based on the dataset used; \(P_\text{max}\) is the maximum number of parameters across all models; and \(w_c\) and \(w_\text{parm}\) are the weights for each criterion (e.g., simplicity, transparency, explainability, number of parameters). 

The interpretability scores for the embedding models have been assigned based on established perceptions in the NLP field. Specifically, CountVectorizer is assigned a score of 0.25, TF-IDF a score of 0.50, and Word2Vec a score of 0.75. Using this methodology, models with lower interpretability exhibit higher interpretability scores compared to more interpretable models. For composite models, these interpretability scores are aggregated from the individual ratings of the constituent models, as follows:
\begin{equation}
  \text{CI Score} = \sum_{i=1}^{n} \text{IS}_i
  \label{eqn:CI_Score}
\end{equation}
\noindent where: \(\text{CI Score}\) is the total interpretability score for the composite model; \(\text{IS}_i\) represents the interpretability score of the \(i\)-th constituent model in the composite model, which could either be an embedding, sentiment analysis or classification model; and \(n\) is the number of constituent models in the composite model. For instance, a composite model that combines TF-IDF review embeddings (score = 0.50) and BERT sentiment scores (score = 1.00), classified using logistic regression (score = 0.22), would be assigned a cumulative interpretability score of 1.72.

\section{Results}

\begin{figure}[!h]
    \centering
    \subfigure[Cell Phones and Accessories (\textbf{CPA})]{\includegraphics[width=0.22\textwidth, height = 1.4in]{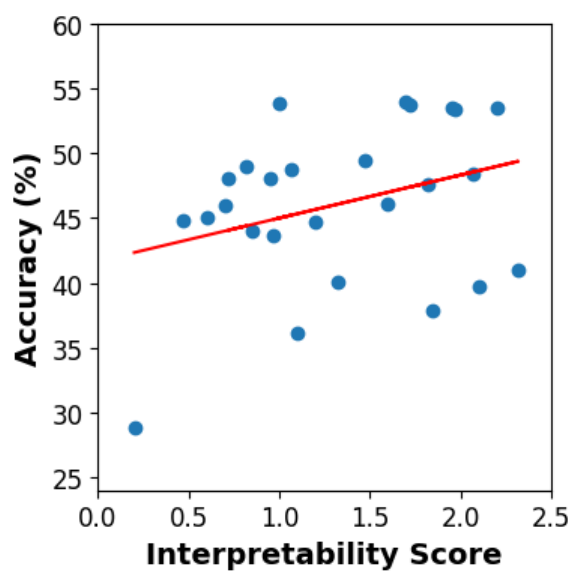}} 
    \subfigure[Office Products (\textbf{OP})]{\includegraphics[width=0.22\textwidth, height = 1.4in]{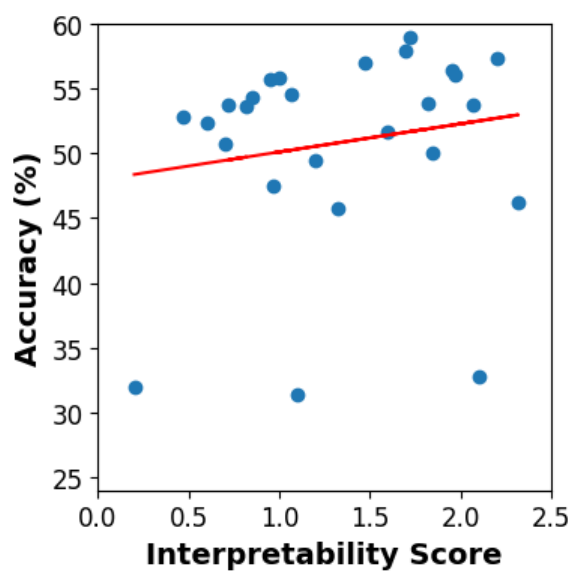}} 
    \subfigure[Automotive (\textbf{AM})]{\includegraphics[width=0.22\textwidth, height = 1.4in]{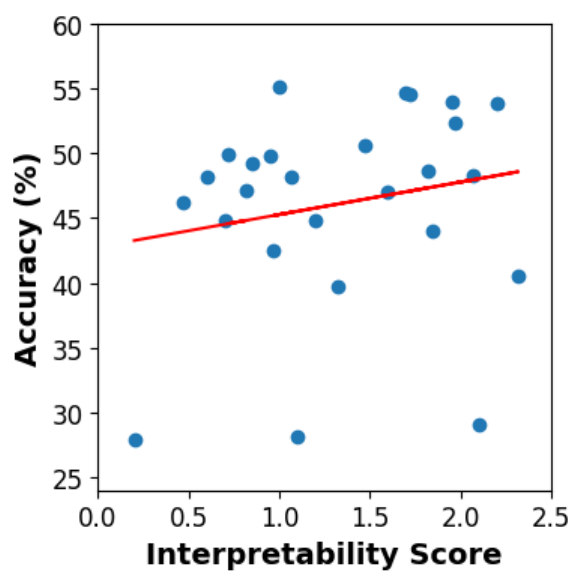}}
    \subfigure[Video Games (\textbf{VG})]{\includegraphics[width=0.22\textwidth, height = 1.4in]{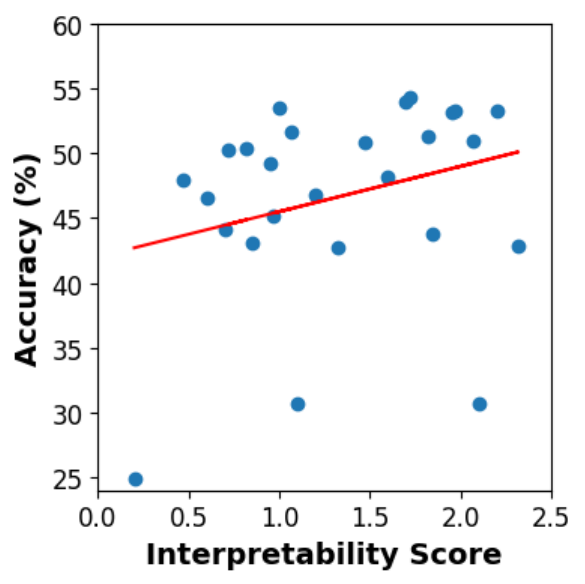}}
    \caption{Trade-off between Interpretability Scores vs Accuracy for 4 datasets: Higher interpretability scores weakly correlate with improved accuracy, but instances exist where highly interpretable models show lower accuracy and less interpretable models perform better.}
    \label{fig:interscores}
\end{figure}

\textit{Accuracy-Interpretability Trade-off:} We compare various model types by classifying numerical ratings based on text reviews. There are a total of 26 distinct composite models with unique interpretability scores for each model type, demonstrating the diverse range of models employed in this study. The interpretability scores are plotted against their corresponding accuracies for each dataset, as illustrated in Figure~\ref{fig:interscores}. Each plot includes a best-fit line to better understand the trends. The analysis reveals that the trend is not very strictly monotonic with several outliers. While higher interpretability scores generally (weakly) correlate with improved accuracies, there are multiple instances where models with high interpretability scores exhibit lower accuracy, and models with lower interpretability scores exhibit higher accuracy. This observation underscores the nuanced relationship of the trade-off between interpretability and performance, indicating that increased interpretability does not consistently translate to enhanced accuracy in all cases. 
    
Based on all four datasets, the models with interpretability scores of 0 (VADER), 1.10 (Word2Vec + NB), and 2.10 (Word2Vec + NB-BS) consistently exhibit poorer performance compared to their more interpretable counterparts. Notably, using the NB classifier with Word2Vec embeddings results in significantly lower accuracies. We hypothesize that the observed trend is due to the NB classifier's assumption of conditional independence among features. Since Word2Vec captures contextual relationships between words, this assumption leads to suboptimal performance because the NB classifier treats word vectors as independent, failing to account for their contextual dependencies.

%(i) Cell Phones and Accessories (\textbf{CPA}); (ii) Office Products (\textbf{OP}); (iii) Automotive (\textbf{AM}); and (iv) Video Games (\textbf{VG}). 

\begin{table*}[h!]
\centering
\small
\caption{Classification of ratings based on reviews using fixed embedding (TF-IDF)}
\label{tab:classification}
\begin{tabular}{c c c c c c c c c c c c}
\hline
\multicolumn{7}{c|}{Highest to lowest interpretability} & \multicolumn{4}{c}{Composite models}
\\ \hline
 & VADER & LR & NB & SVM & NN & BERT & LR-BS & NB-BS & SVM-BS & NN-BS\\ 
 \hline
CPA &  28.82\% & 48.07\% & 44.00\% & 48.03\% & 48.80\% & \textbf{53.82\%} & 53.70\% & 37.87\% & 53.47\% & 48.40\%\\

OP & 31.93\% & 53.80\% & 54.30\% & 55.70\% & 54.57\% & 55.79\% & \textbf{58.97\%} & 50.03\% & 56.40\% & 53.80\%\\

AM & 27.92\% & 49.93\% & 49.26\% & 49.77\% & 48.13\% & \textbf{55.13\%} & 54.60\% & 44.00\% & 53.93\% & 48.27\%\\

VG & 24.86\% & 50.23\% & 43.07\% & 49.23\% & 51.63\% & 53.53\% & \textbf{54.33\%} & 43.80\% & 53.16\% & 50.93\%\\
\hline
\end{tabular}
\end{table*}

Table \ref{tab:classification} presents the accuracy results for classification models while keeping the embedding fixed (TF-IDF) across four datasets (others in Appendix). Accuracies typically range from 40\% to 60\%, with only modest differences observed. Generally, as model interpretability decreases, accuracy tends to improve. This implies that models with lower interpretability, often characterized by higher complexity, can achieve better performance metrics for specific learning tasks. In the case of composite models, the integration of BERT sentiment scores as an additional input generally contributes to an enhancement in accuracy. This is indicative of the added value that more complex models can capture intricate data relationships and enhance the learning if the test data shares similar characteristics to the training data. 

However, this pattern is not universal across all model types. Notably, combining NB with BERT sentiment scores results in reduced accuracy. Moreover, as model complexity increases for composite models, there is a tendency for accuracies to decline. This indicates that while more complex black-box models might offer improved performance in some instances, their effectiveness does not consistently correlate with increased complexity. The performance of composite black-box models does not universally improve with higher complexity, therefore highlighting the need for a nuanced approach when selecting models and balancing between interpretability and performance.

\noindent \textit{Correlation Analysis:} To compare the most interpretable (VADER) and least interpretable (BERT) models in our study, we analyze the correlation between sentiment scores derived from text reviews and the corresponding ratings. Pearson’s correlation~\citep{pearson1896mathematical} ranges from -1 to 1, where 1 indicates a perfect positive correlation, 0 denotes no correlation, and -1 represents a perfect negative correlation. Table \ref{tab:corr_sent} presents the correlations between ratings and sentiment scores from both BERT and VADER across the four datasets. The results show a marked difference in the correlation strengths. VADER, as a highly interpretable model, exhibits moderate positive correlations with ratings, ranging from 0.3 to 0.5. In contrast, BERT, a less interpretable model, shows a significantly stronger positive correlation with ratings, ranging from 0.7 to 1. This discrepancy is anticipated, as BERT sentiment models are trained on product reviews, whereas VADER sentiment models are optimized for sentiments expressed in social media contexts. These findings highlight the trade-off between interpretability and accuracy in model selection, where more interpretable models show moderate correlations, while less interpretable models provide stronger predictive accuracy.

\begin{table}[h]
\caption[Correlation between sentiment scores and ratings]{Correlation between sentiment scores and ratings}
\begin{center}
\small
\begin{tabular}{ c c c c }
\hline
Dataset & VADER & BERT\\
\hline
Cell Phone and Accessories (CPA) & 0.4328 & 0.7675\\
Office Products (OP) & 0.5098 & \textbf{0.8202}\\
Automotive (AM) & 0.4165 & 0.7871\\
Video Games (VG) & 0.3879 & 0.7653\\
\hline
\end{tabular}
\end{center}
\label{tab:corr_sent}
\end{table}

\begin{figure}[h]
    \centering
    \subfigure[OP - VADER]{\includegraphics[width=0.22\textwidth, height = 1.2in]{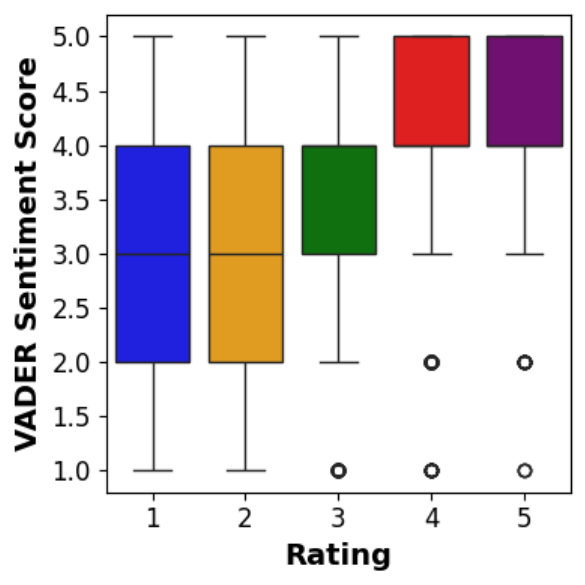}}
    \hfill
    \subfigure[OP - BERT]{\includegraphics[width=0.22\textwidth, height = 1.2in]{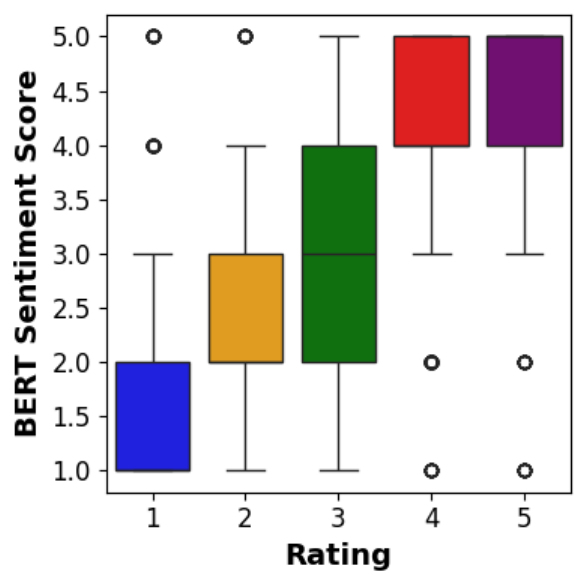}} 
    \hfill
    \subfigure[CPA - VADER]{\includegraphics[width=0.22\textwidth, height = 1.2in]{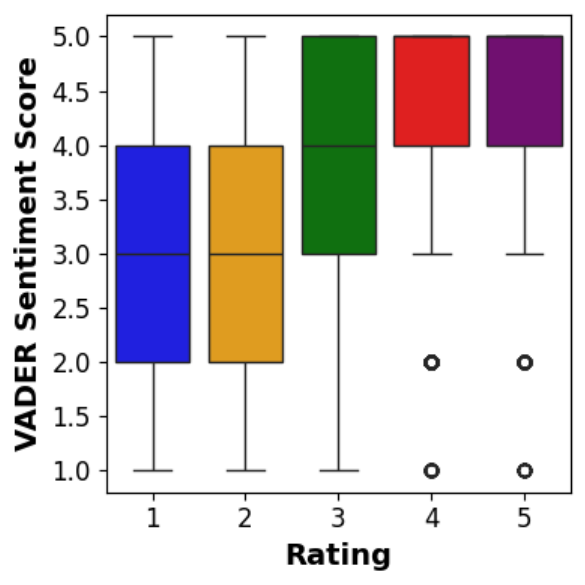}}
    \hfill
    \subfigure[CPA - BERT]{\includegraphics[width=0.22\textwidth, height = 1.2in]{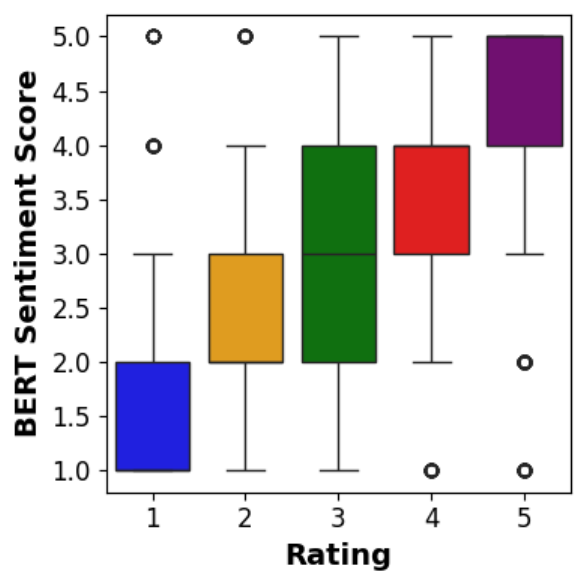}}
    \caption{Box-plots of sentiment scores vs. ratings in CPA and OP datasets.}
    \label{fig:boxplots}
\end{figure}

The relationship between the two sentiment analysis models is further examined through box plots, as illustrated in Figure~\ref{fig:boxplots}. Box-plots of both VADER and BERT sentiment scores are plotted against ratings for two datasets: CPA and OP. It can be observed that VADER sentiment scores exhibit greater skewness compared to BERT sentiment scores. For example, in the CPA dataset, the interquartile range for VADER sentiment scores associated with 1-star reviews spans from approximately 2 to 4. In contrast, the interquartile range for BERT sentiment scores for 1-star reviews in the same dataset ranges from approximately 1 to 2. Additionally, the presence of numerous outliers underscores the limitations of relying exclusively on sentiment analysis tools for this particular use case. These visualizations indicate that BERT sentiment scores provide a more accurate correlation with ratings compared to VADER sentiment scores.

\section{Conclusion}
This work studies the trade-offs between interpretability and performance across various machine learning models, specifically focusing on the NLP task of predicting product ratings from textual reviews. Our comparative analysis of both black-box and interpretable models, coupled with the introduction of a quantitative measure for interpretability, provides valuable insights into the interplay between these two critical aspects of model selection. The findings reveal a general trend where model accuracy tends to improve as interpretability decreases. Yet this relationship is not universally applicable. Composite black-box models do not necessarily outperform combinations of interpretable and black-box models. There are notable instances where more interpretable models perform better. The limitations of this study are discussed in detail in the Appendix.

\noindent \textbf{Future Work:} We aim to develop an advanced framework for ranking models based on interpretability, replacing expert assessments with analytical factors such as feature importance scores and decision boundaries. Leveraging advances in explainable AI, techniques such as LIME by \citet{ribeiro2016should} may enable deeper insights into the decision-making processes of models, enhancing the understanding of both model accuracy and interpretability.

\begin{table*}[h!]
\centering
\caption{Classification of ratings based on reviews using fixed embedding (CountVectorizer)}
\label{tab:classification_CV}
\begin{tabular}{c c c c c c c c c c c c}
\hline
\multicolumn{7}{c|}{Highest to lowest interpretability} & \multicolumn{4}{c}{Composite models}
\\ \hline
 & VADER & LR & NB & SVM & NN & BERT & LR-BS & NB-BS & SVM-BS & NN-BS\\ 
 \hline
CPA &  28.82\% & 44.77\% & 45.03\% & 46.00\% & 49.00\% & 53.82\% & 49.40\% & 46.13\% & 54.00\% & 47.66\%\\

OP & 31.93\% & 52.87\% & 52.30\% & 50.73\% & 53.60\% & 55.79\% & 57.03\% & 51.70\% & 57.90\% & 53.87\%\\

AM & 27.92\% & 46.23\% & 48.13\% & 44.87\% & 47.13\% & 55.13\% & 50.63\% & 47.03\% & 54.70\% & 48.60\%\\

VG & 24.86\% & 47.97\% & 46.60\% & 44.10\% & 50.43\% & 53.53\% & 50.83\% & 48.20\% & 54.03\% & 51.27\%\\
\hline
\end{tabular}
\end{table*}

\begin{table*}
\centering\caption{Classification of ratings based on reviews using fixed embedding (Word2Vec)}
\label{tab:classification_W2V}
\begin{tabular}{c c c c c c c c c c c c}
\hline
\multicolumn{7}{c|}{Highest to lowest interpretability} & \multicolumn{4}{c}{Composite models}
\\ \hline
 & VADER & LR & NB & SVM & NN & BERT & LR-BS & NB-BS & SVM-BS & NN-BS\\ 
 \hline
CPA &  28.82\% & 43.67\% & 36.10\% & 44.67\% & 40.10\% & 53.82\% & 53.40\% & 39.73\% & 53.46\% & 41.03\%\\

OP & 31.93\% & 47.53\% & 31.33\% & 49.43\% & 45.80\% & 55.79\% & 56.07\% & 32.80\% & 57.30\% & 46.23\%\\

AM & 27.92\% & 42.47\% & 28.10\% & 44.77\% & 39.67\% & 55.13\% & 52.30\% & 29.10\% & 53.90\% & 40.56\%\\

VG & 24.86\% & 45.13\% & 30.70\% & 46.83\% & 42.70\% & 53.53\% & 53.23\% & 30.73\% & 53.27\% & 42.80\%\\
\hline
\end{tabular}
\end{table*}

\section{Appendix}
\subsection{Additional Experiments}

Similar to Table \ref{tab:classification}, which shows classification accuracies using TF-IDF embeddings, Tables \ref{tab:classification_CV} and \ref{tab:classification_W2V} present classification accuracies for CountVectorizer and Word2Vec embeddings, respectively, across four datasets. These tables exhibit a similar trend: generally, as model interpretability decreases, accuracy tends to improve. For composite models, incorporating BERT sentiment scores as an additional input usually leads to better accuracy. However, when it comes to composite models, increasing model complexity does not show a consistent pattern in how interpretability affects performance.

\begin{figure}[h!]
    \centering    \subfigure{\includegraphics[width=0.14\textwidth, height = 1.1in]{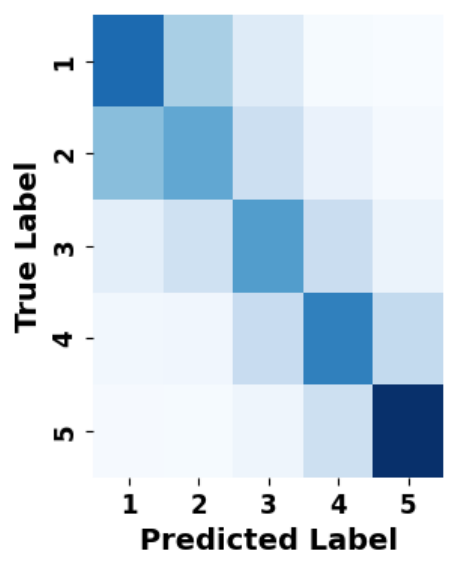}}\hfill
    \subfigure{\includegraphics[width=0.14\textwidth, height = 1.1in]{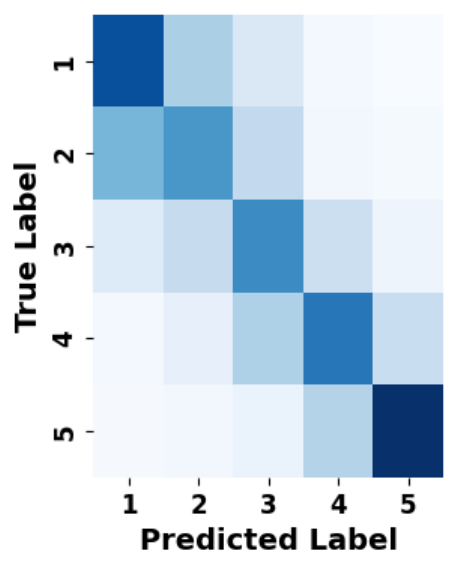}}\hfill
    \subfigure{\includegraphics[width=0.14\textwidth, height = 1.1in]{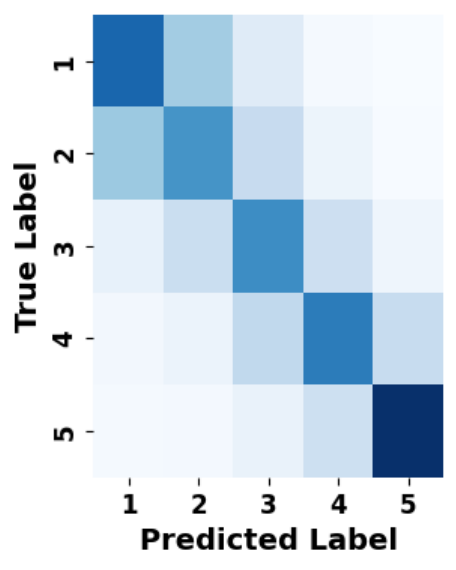}}\hfill
    \subfigure{\includegraphics[width=0.14\textwidth, height = 1.1in]{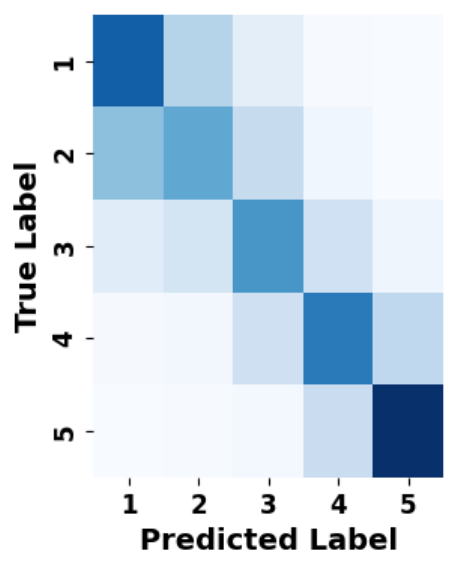}}\hfill
    \subfigure{\includegraphics[width=0.14\textwidth, height = 1.1in]{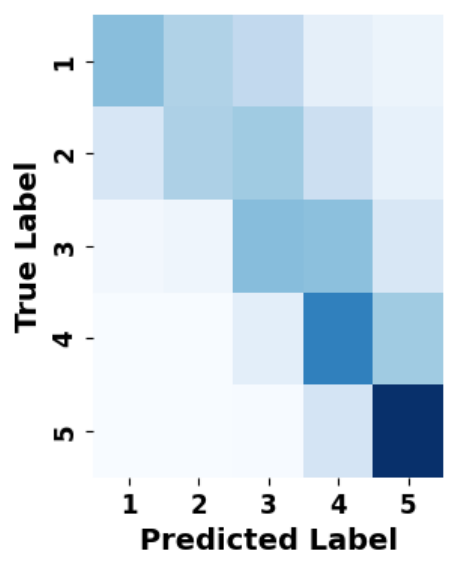}}\hfill
    \subfigure{\includegraphics[width=0.14\textwidth, height = 1.1in]{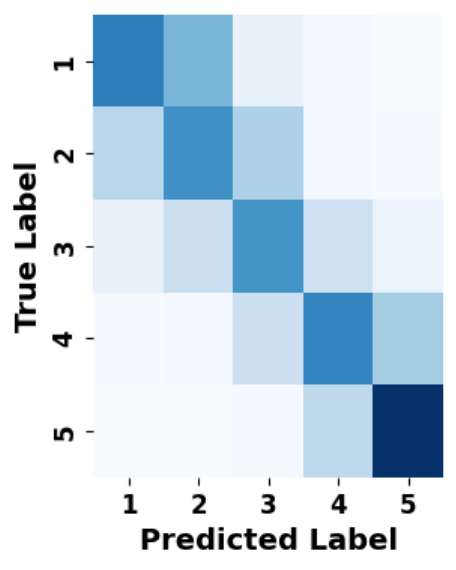}}\hfill
    \subfigure{\includegraphics[width=0.45\textwidth, height = 0.2in]{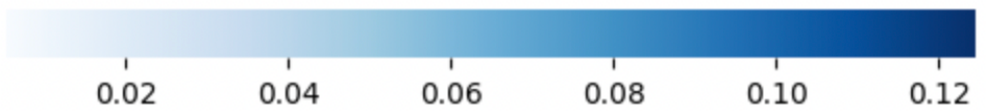}}\hfill
    \caption{Confusion matrices for LR, NB, SVM, LR-BS, NB-BS, and SVM-BS in OP dataset.}
    \label{fig:cfmatrices}
\end{figure}

Confusion matrices were computed for the following classification models while fixing the embedding as TF-IDF: LR, NB, SVM, LR-BS, NB-BS, SVM-BS in the OP dataset, as shown in Figure~\ref{fig:cfmatrices}. The matrices illustrate that diagonal entries indicate the frequency of instances where the predicted label matches the true label. Ratings are predominantly classified into their corresponding classes or adjacent rating classes; for example, reviews with a 2-star rating are frequently classified into 1-star, 2-star, or 3-star categories. Although the matrices show minor differences among the models, it is evident that the classifiers tend to predict within their assigned classes or nearby rating classes.

\subsection{Limitations}
This study is specifically designed to evaluate the performance of models for predicting product ratings based on textual reviews. Consequently, the results may not be generalizable to other application domains or datasets with different characteristics. Furthermore, our analysis is restricted to particular models, including VADER and BERT for sentiment analysis, along with selected classification algorithms. This focus may exclude other potentially relevant models or methodologies, which may behave differently. Additionally, the proposed quantitative score relies on expert assessments, necessitating expert participation for future analyses of other models and algorithms.

\section{Acknowledgments}
We would like to thank the domain experts who generously participated in our survey and provided valuable insights into the interpretability of machine learning models.

\bibliography{aaai25}

\end{document}